\documentclass[runningheads]{llncs}
\usepackage[T1]{fontenc}
\usepackage{graphicx}
\usepackage{amsmath}
\usepackage{amssymb}
\usepackage{orcidlink}
\usepackage{makecell}
\usepackage{hyperref}
\usepackage{url}
\begin{document}
\title{Enhanced Sparse Point Cloud Data Processing for Privacy-aware Human Action Recognition}

\titlerunning{Enhanced Sparse Point Cloud Data Processing}
%
\author{Maimunatu Tunau\inst{1} \orcidlink{0009-0001-7255-4577} \and
Vincent Gbouna Zakka\inst{1} \orcidlink{0000-0001-9360-0985} \and
Zhuangzhuang Dai\inst{1} \orcidlink{0000-0002-6098-115X}}

\authorrunning{M. Tunau et al.}

\institute{Department of Applied Artificial Intelligence and Robotics, Aston University, Birmingham, UK}
%
%
\maketitle              
\begin{abstract}
Human Action Recognition (HAR) plays a crucial role in healthcare, fitness tracking, and ambient assisted living technologies. While traditional vision-based HAR systems are effective, they pose privacy concerns. mmWave radar sensors offer a privacy-preserving alternative but present challenges due to the sparse and noisy nature of their point cloud data. In the literature, three primary data processing methods—Density-Based Spatial Clustering of Applications with Noise (DBSCAN), the Hungarian Algorithm, and Kalman Filtering—have been widely used to improve the quality and continuity of radar data. However, a comprehensive evaluation of these methods, both individually and in combination, remains lacking. This paper addresses that gap by conducting a detailed performance analysis of the three methods using the MiliPoint dataset. We evaluate each method individually, all possible pairwise combinations, and the combination of all three, assessing both recognition accuracy and computational cost. Furthermore, we propose targeted enhancements to the individual methods aimed at improving accuracy. Our results provide crucial insights into the strengths and trade-offs of each method and their integrations, guiding future work on mmWave-based HAR systems.Our source code is made publicly available at \footnote{\url{https://github.com/Maimunatunau/Human-Action-Recognition-HAR-using-mmWave-Radar}.}
\keywords{Human action recognition \and  Point cloud data processing \and Privacy-aware human action recognition}
\end{abstract}

\section{Introduction}
In the era of intelligent environments and pervasive computing, the ability for systems to perceive and respond to human actions in real time has become fundamental to the development of smart technologies. From adaptive lighting and automated security in smart homes to fall detection systems for elderly care, Human Action Recognition (HAR) plays a crucial role in enabling responsive, context-aware applications. Traditionally, HAR has relied on camera-based systems due to their rich visual information. However, these systems come with some limitations, including privacy concerns, dependence on good lighting conditions, and high computational demands~\cite{Karim2024}.

Millimeter-wave (mmWave) radar has emerged as a promising alternative for HAR, offering privacy-preserving, lighting-invariant sensing capabilities. Operating in high-frequency bands, mmWave radar enables robust motion detection even in visually obscured environments~\cite{Huang2023}. When combined with deep learning models, this sensing modality has achieved considerable success in recognizing human actions~\cite{Zeng2020}. Still, challenges remain—particularly in dealing with the sparse, noisy point cloud data generated by radar sensors, which can hinder both feature extraction and accurate classification~\cite{Cui2021}.

To address these challenges, several preprocessing approaches have been introduced in recent studies, focusing on clustering, tracking, and temporal association. For example,~\cite{Zhao2019} introduced the mID pipeline, combining DBSCAN clustering, the Hungarian Algorithm, and a Kalman Filter, but their work lacked systematic tuning and degraded with sparse input. Other studies explored deep learning enhancements to Kalman filtering~\cite{Pegoraro2021}, and~\cite{Dang2024} emphasized fast clustering and association strategies. While each of these approaches contributes valuable insights, they are often evaluated in isolation. A unified and systematic evaluation of these methods—and their combinations—is yet to be explored.

This paper addresses that gap by presenting a comprehensive evaluation of three widely used radar point cloud preprocessing techniques: DBSCAN, the Hungarian Algorithm, and Kalman Filtering. Our focus is on analyzing how each method performs individually, how they interact in pairs, and how they behave when combined—all using the MiliPoint dataset as a case study. We assess both recognition accuracy and computational efficiency to better understand the trade-offs involved in using each method or combination. In addition, we propose practical enhancements to the individual techniques to boost performance, particularly in noisy environments. By conducting this systematic evaluation, we aim to inform the design of more effective mmWave radar-based HAR systems and provide guidance for selecting or combining the preprocessing methods. 
\begin{figure}[h!]
    \centering
    \includegraphics[width=0.7\linewidth]{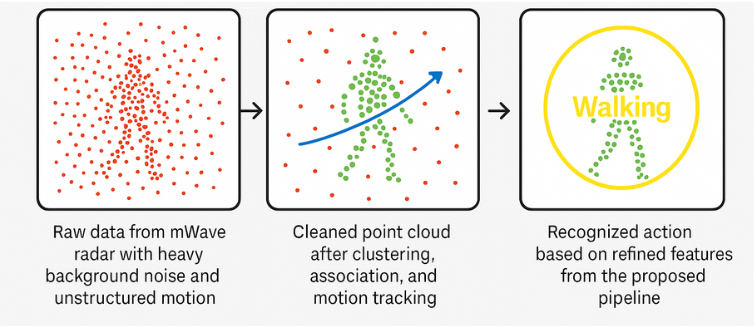} 
    \caption{Challenges and opportunities in mmWave radar-based Human Action Recognition. Left: Raw mmWave data is noisy, sparse, and lacks meaningful structure. Middle: Integration and application of data processing pipeline. Right: Our proposed pipeline (DBSCAN, Hungarian Algorithm, Kalman Filter) mitigates noise, preserves temporal coherence, and enables reliable action recognition.}
    \label{fig:intro}
\end{figure}
\section{Related Work}
\subsection{Evolution of Human Action Recognition (HAR)}
HAR has undergone substantial development over the past two decades, driven by advancements in artificial intelligence, sensor technologies, and computational capabilities. Early HAR systems predominantly relied on vision-based techniques such as Optical Flow and Histogram of Oriented Gradients (HOG), which extracted motion and shape descriptors from video sequences~\cite{Aggarwal2011}. While these approaches achieved moderate success in controlled environments, they were highly sensitive to changes in lighting, occlusions, and raised concerns regarding user privacy.

The introduction of machine learning algorithms, for instance Support Vector Machines (SVMs), marked a step forward in HAR by enabling supervised classification of complex motion patterns~\cite{Wahyu2022}. However, these approaches required extensive feature engineering, and often struggled with scalability in multi-class action recognition settings. More recently, deep learning particularly Convolutional Neural Networks (CNNs) and Recurrent Neural Networks (RNNs) has largely replaced the classical models by automating feature extraction and improving recognition performance across diverse datasets~\cite{Zeng2020,RETCN}.

Despite these improvements, vision-based systems remain constrained by their dependency on ambient lighting and potential invasiveness. As a result, recent methods has increasingly turned to alternative sensing modalities, such as inertial sensors, infrared systems, and, more recently, millimetre-wave (mmWave) radar. Radar-based HAR offers several advantages, including lighting invariance, low-power operation, and preservation of personal privacy. Nevertheless, HAR using mmWave radar faces unique challenges related to data sparsity, and noise~\cite{Cui2021}.

\subsection{Data Processing in mmWave Radar-Based HAR}
Millimetre-wave radar, operating in the 30–300,GHz range, enables high-resolution spatial sensing through wide bandwidths and fine Doppler measurements~\cite{Appleby2007}. This makes it suitable for detecting subtle human movements even in poor lighting conditions. However, radar signals often contain substantial noise and non-target reflections, necessitating data preprocessing pipelines to ensure reliable action recognition.

To mitigate noise and improve tracking, recent methods have employed various data processing techniques. Density-Based Spatial Clustering of Applications with Noise (DBSCAN) is frequently used for its ability to identify arbitrarily clusters and filter out outliers without requiring prior knowledge of the number of clusters~\cite{Ahmed2016}. Compared to traditional clustering methods like K-means, DBSCAN is more resilient to noisy and irregular radar point cloud data~\cite{Andriyani2022}.

For temporal consistency and object tracking, the Hungarian Algorithm (HA) has been widely adopted for optimal frame-to-frame association of clusters or detected centroids~\cite{Kuhn1955}. HA provides computationally efficient solutions to assignment problems and demonstrates improved performance over alternatives such as the Auction Algorithm in large-scale scenarios~\cite{Arunachalam2020}. Other methods have used Kalman Filtering (KF) to predict and update object states, effectively smoothing trajectories and handling partial observations in dynamic scenes~\cite{Welch1995}. In HAR applications involving near-linear human motion, the standard Kalman Filter has proven to be both effective and computationally lightweight~\cite{Campestrini2016}.

Despite the application of these individual techniques, there has been little systematic analysis of their individual performance or how they behave in different combinations. This study therefore presents an evaluation of these methods, assessing both their recognition accuracy and computational cost, and further propose enhancements to each technique to improve performance.

\section{Methodology}
\begin{figure}[h!]
    \centering
    \includegraphics[width=0.6\linewidth]{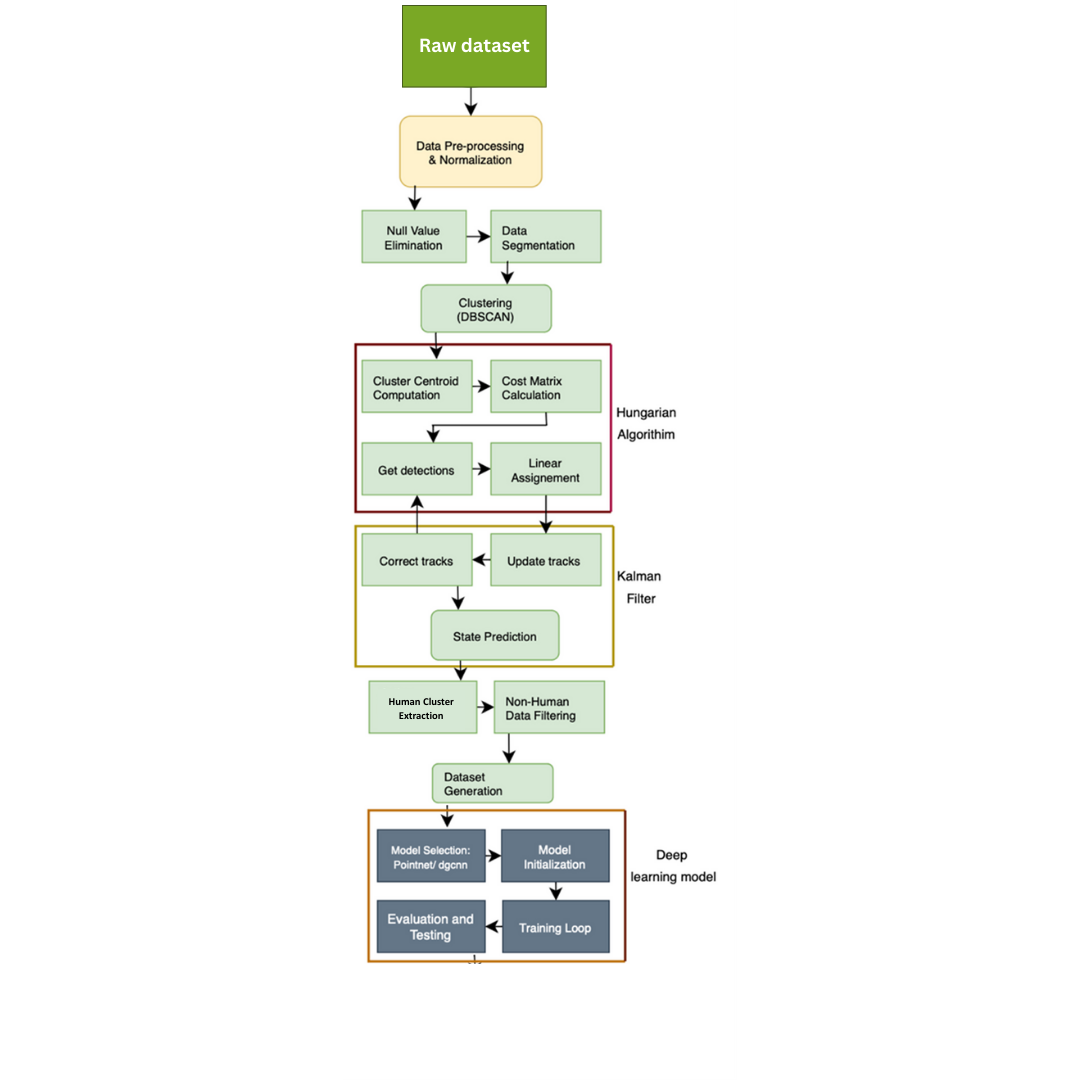} 
    \vspace{-20pt} 
    \caption{Flowchart of the proposed Human Action Recognition pipeline.}
    \label{fig:flowchart}
\end{figure}

The framework of the proposed data processing pipeline is designed to progressively refine the raw point cloud data from noise filtering and segmentation through to clustering, temporal association, and trajectory prediction—ensuring that the strengths of each method contribute to a robust overall system, this flow is seen in \ref{fig:flowchart}.
\subsection{Dataset and Preparation}
We used the MiliPoint dataset~\cite{Cui2023}, a structured point-cloud dataset for HAR recorded using mmWave radar sensor. It comprise 545,000 frames collected from 11 participants performing 49 distinct actions. Each frame consists of \((x,y,z)\) coordinates. Following the approach in~\cite{Cui2023}, temporal data were stacked up to 1100 points per frame. The dataset was partitioned into training (80\%), validation (10\%), and testing (10\%).
\subsection{Segmentation}
\label{sec:segmentation}
To ensure robust clustering and preserve temporal continuity, each raw frame is subdivided into multiple contiguous segments before any clustering or tracking. We evaluated three splitting strategies being 2, 5, and 10 segments per frame and found that dividing each frame into five equal segments provided the best balance between noise removal and data retention. Specifically, each raw radar frame is partitioned into five segments, each containing 220 consecutive (x,y,z) points. By dividing frames into smaller segments, we reduce within‑segment variance and improve downstream clustering consistency.
\subsubsection{Null-Value Removal}
Within each segment, points near the origin (zero‑padding) are discarded by applying an L2-norm threshold:
\begin{equation}
(p_x^2 + p_y^2 + p_z^2) > \text{threshold}^2,\quad \text{with } \text{threshold}=0.001.
\end{equation}
This step removes the zero-padding, preventing DBSCAN from misinterpreting filler points as meaningful data. Following segmentation and L2-norm–based null-value removal, we observed that the five-segment configuration consistently retained the optimal number of valid points per segment for effective clustering. On average, each of the five segments retained approximately 63 non-zero points after thresholding, compared to ~132 points for the 2-segment split and ~32 points for the 10-segment split. The 2-segment configuration retained a higher number of points but introduced significant intra-segment variation, which negatively impacted DBSCAN’s ability to form tight, coherent clusters. In contrast, the 10-segment configuration overly fragmented the data, often resulting in segments with insufficient points for meaningful clustering. The five-segment approach struck the best balance, preserving temporal granularity while maintaining sufficient point density. This configuration reduced noise from zero-padding while avoiding data sparsity, leading to more stable clustering and improved downstream performance. Consequently, the five-segment split was selected for all subsequent stages of the pipeline. 
\subsection{Clustering and Tracking Pipeline}
\subsubsection{DBSCAN for Noise Reduction and Cluster Identification}
We conducted parameter tunning to select the most efficient parameters for DBSCAN. After parameter tunning, the following parameters were selected: \(\epsilon=0.4\) and \(\text{min\_samples}=6\). Additionally, we introduced a vertical weighting factor \(\alpha=0.25\) in the Euclidean metric. This adjustment de-emphasizes the \(z\)-axis, aligning the clustering process with the predominantly horizontal human motion captured by the radar. The clusters detected by DBSCAN provide refined spatial representations that not only filter out environmental noise but also serve as the basis for subsequent data association.
\subsubsection{Data Association Using the Hungarian Algorithm (HA)}
To construct continuous motion trajectories, we compute the centroids of the clusters as follows:
\begin{equation}
    Centroid=\left(\frac{1}{n}\sum_{i=1}^{n} x_i,\, \frac{1}{n}\sum_{i=1}^{n} y_i,\, \frac{1}{n}\sum_{i=1}^{n} z_i\right).
    \label{centroid}
\end{equation}
A cost matrix based on the Euclidean distances between centroids across adjacent segments is then formed. The HA uses this matrix to optimally match clusters over time across segments, ensuring a coherent assignment of tracks across frames. This step is important because accurate inter-segment association prepares the data for reliable trajectory prediction.
\subsubsection{Trajectory Prediction with Kalman Filtering (KF)}
\label{sec:kalman_filtering}
With the associated clusters, the Kalman Filter is then applied to predict and correct the trajectory of the associated clusters using the state vector equation \ref{state} and \text{and observation vector} equation \ref{observation}of the detected human cluster. The state vector is defined as
\begin{equation}
     X=[\,x,\;y,\;v_x,\;v_y\,]^T,
     \label{state}
\end{equation}
and the observation vector is
\begin{equation}
    z=[x,\;y]^T.
    \label{observation}
\end{equation}
To optimize the filter’s performance, we tuned the noise covariance parameters \(Q\), \(R\), and the state covariance \(P\) via Bayesian Optimisation (BO)~\cite{Chen2012} on a sample of frames to prevent bias across frames. The BO process iteratively evaluated parameter settings on multiple sample frames to minimize the prediction error. By iteration 15, an optimal configuration was obtained with \(Q = 29.41\), \(R = 0.081\), and \(P = 14.64\), achieving a minimal prediction error of \(-0.001001\). Additionally, an Euclidean distance threshold of 2.0 ensures that only closely matched clusters are tracked, reducing false associations. This filtering stage compensates for residual noise and provides smooth trajectory estimates, leveraging the reliable associations formed by the HA.

\subsection{Human Cluster Selection and Final Evaluation}
After KF updates, the centroids of the predicted tracks are compared against ground-truth medians of the human keypoints. The prediction accuracy is quantitatively assessed via RMSE:
\begin{equation}
    \text{RMSE} = \sqrt{\frac{1}{N} \sum_{i=1}^{N} ( \text{predicted position}_i - \text{actual position}_i )^2}.
\end{equation}
If one track dominates both the RMSE and the spatial proximity measures, additional logic considers the second-best track to ensure the integrity of the detected human cluster. Ultimately, non-selected clusters are zeroed out, leaving only the primary track that best represents the human cluster. The integration of the different algorithms allows each method to build on the preceding steps: DBSCAN effectively reduces noise and segments the data, HA provides a reliable temporal association across segments, and KF refines the motion estimates by predicting and correcting trajectories.

\section{Results}
\label{sec:evaluation}
In this section, we assess the performance of the proposed data processing pipeline and discuss its implications for human activity recognition (HAR). We evaluate each stage of the pipeline from segmentation and null-value removal through clustering, inter-frame association, and trajectory prediction. Secondly, we compare the performance of each method, their pairwise combinations, and the combination of all three, evaluating both action recognition accuracy and computational cost. And finally, we compare the performance against baseline model architectures.

\subsection{Preliminary Evaluation}
\subsubsection{DBSCAN Clustering}
\begin{figure}[h!]
    \centering
    \includegraphics[width=\textwidth]{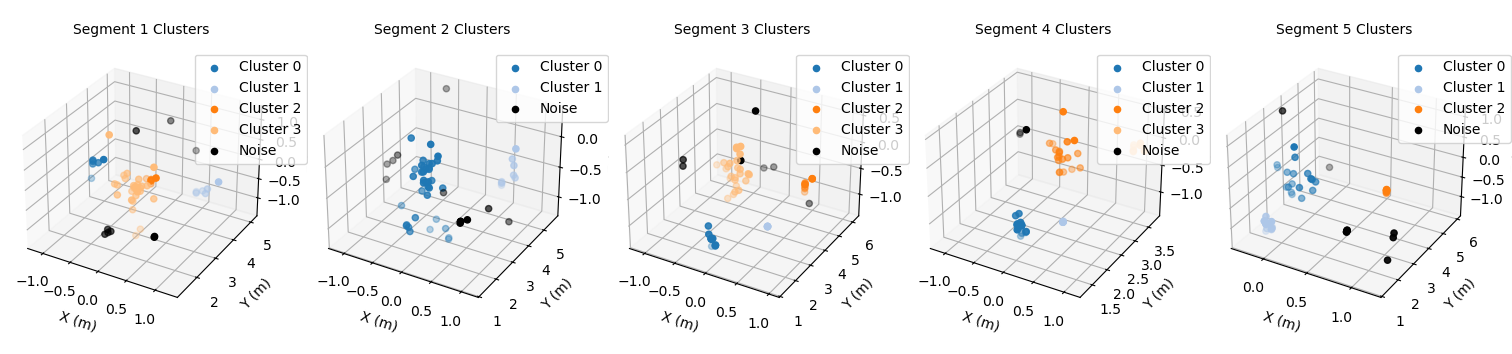}
    \caption{DBSCAN clustering results. Different colors represent distinct clusters; black points indicate noise.}
    \label{fig:DBSCANClusteringResult}
\end{figure}

DBSCAN was applied independently to each of the five segments as described in section~\ref{sec:segmentation} to identify spatial clusters corresponding to human motion. Our experiments revealed that some segments produced well-defined clusters with low noise, while others showed higher variability. Figure~\ref{fig:DBSCANClusteringResult} illustrates the clustering outcomes, where distinct colors denote separate clusters and black points indicate noise.
\subsubsection{Data Association via the Hungarian Algorithm}
\begin{figure}[h!]
    \centering
    \includegraphics[width=0.6\textwidth]{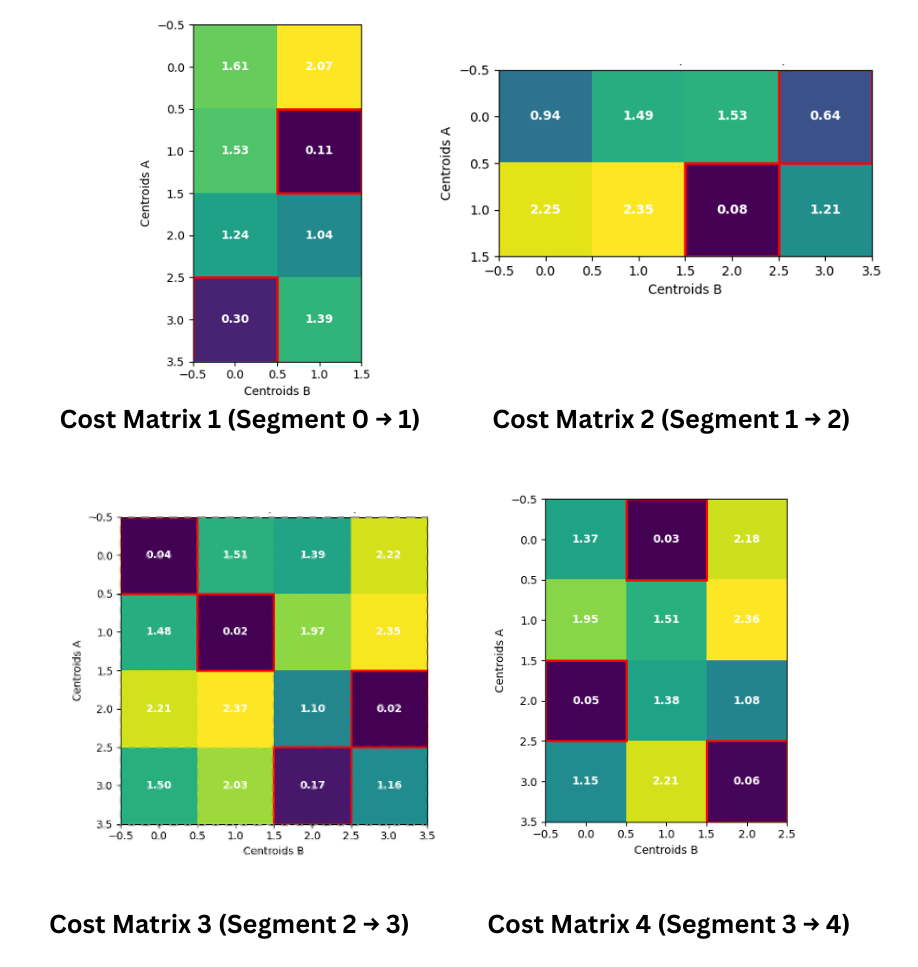}
    \caption{Cost matrices for consecutive segments. Red-highlighted entries denote minimal-cost assignments.}
    \label{fig:CostMatrix}
\end{figure}

\begin{figure}[h!]
    \centering
    \includegraphics[width=1\textwidth]{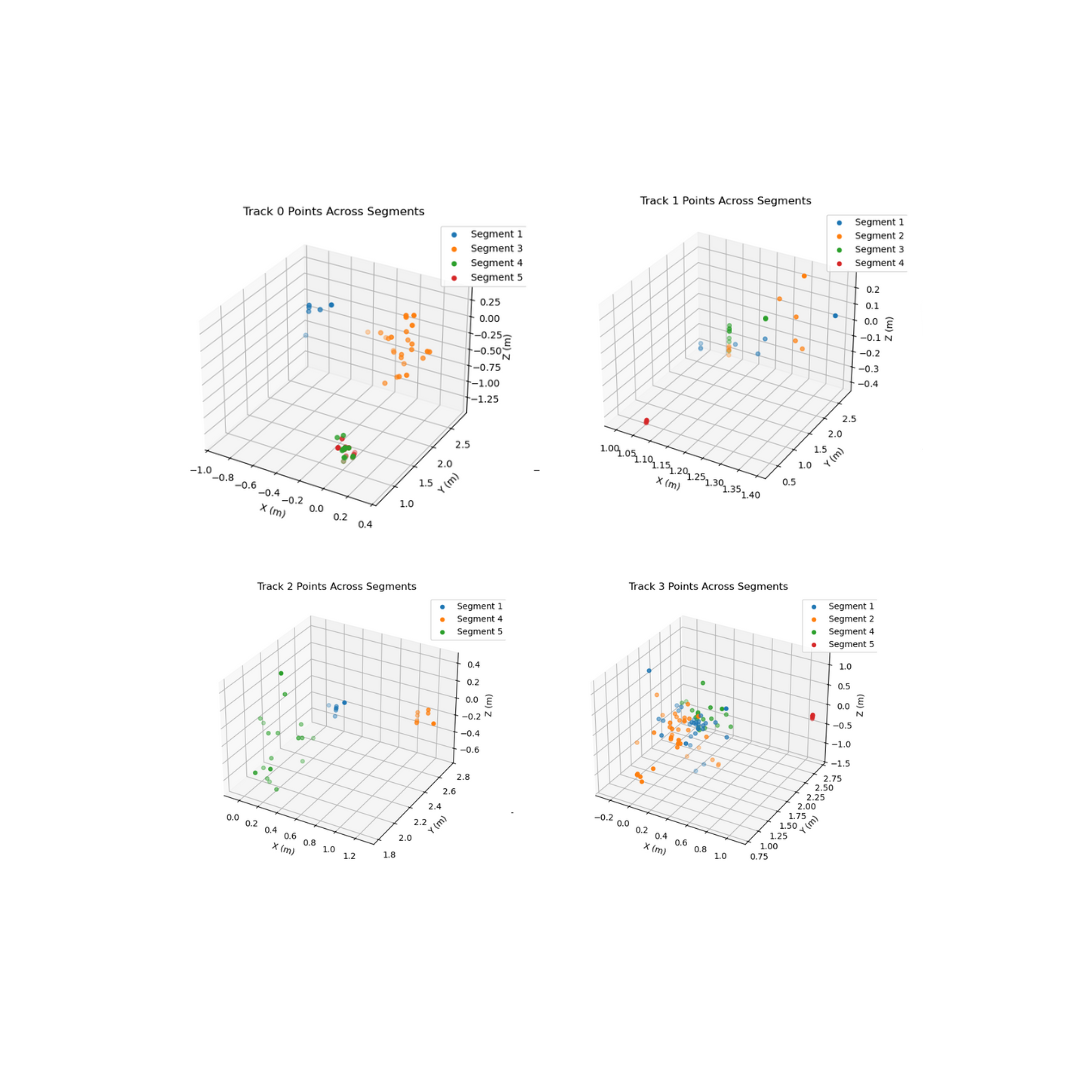}
    \vspace{-60pt} 
    \caption{Visual representation of the four formed tracks. Each colored point corresponds to a distinct segment.}
    \label{fig:tracks}
\end{figure}

To establish temporal continuity across segments, we computed a cost matrix of Euclidean distances between the centroids of clusters in consecutive segments. 
The HA then optimally linked these clusters to form tracks across the entire frame. Our results yielded four tracks, with Tracks~0 and Track~3 showing robust consistency over multiple segments, whereas Tracks~1 and 2, though covering fewer segments, still maintained coherent motion patterns. Figure~\ref{fig:CostMatrix} illustrates representative cost matrices, and Figure~\ref{fig:tracks} visualizes the formed tracks.
\subsubsection{Kalman Filter}
\begin{figure}[tbh!]
    \centering
    \includegraphics[width=0.8\textwidth]{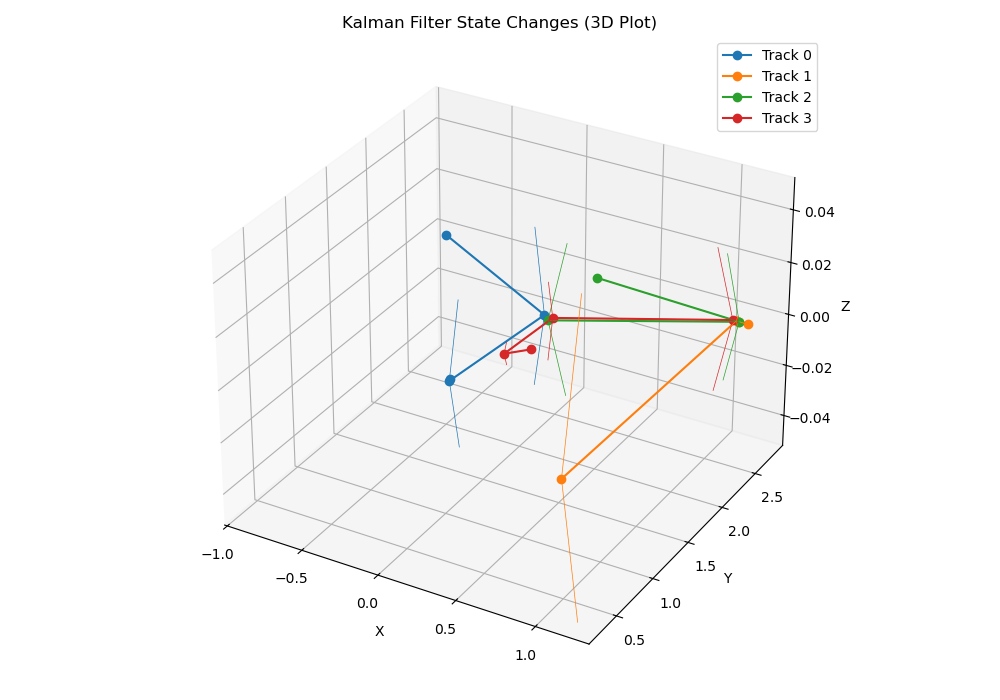}
    \caption{Kalman filter state changes across four tracks. Each colored line indicates transitions over consecutive frames.}
    \label{fig:kalman_state_change}
\end{figure}

\begin{figure}[h!]
    \centering
    \includegraphics[width=0.5\textwidth]{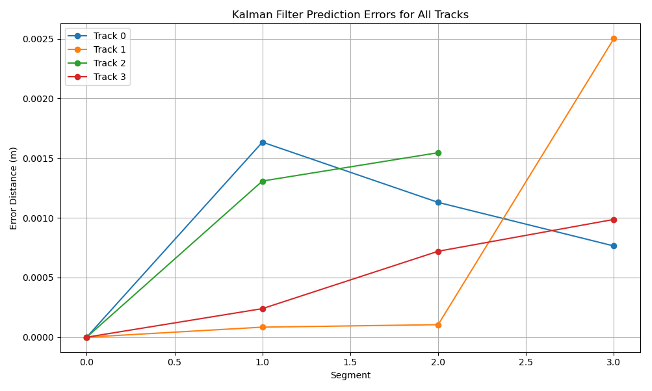}
    \caption{Prediction error in the \(xy\)-plane for all tracks. The vertical axis represents error distance, and the horizontal axis represents segments.}
    \label{fig:prediction_error}
\end{figure}

After applying Kalman filter as decribed in section~\ref{sec:kalman_filtering}, all tracks stabilized with Track~0 showing a state change of approximately 1.19 at iteration 2, and near-zero change (around 0.02) by iteration 4. In contrast, Tracks~1 and 2 experienced larger transitions, indicating higher noise, while Track~3 stabilized rapidly despite a higher point density. Figures~\ref{fig:kalman_state_change} and~\ref{fig:prediction_error} illustrate the state changes and prediction errors, respectively.
\subsection{Human Cluster Analysis}
We compared the final clusters to ground-truth keypoints, leveraging RMSE and median distance to quantify alignment. Table \ref{tab:rmse_distance} summarizes RMSE and median distance for each track using one sample. Track3 showed the lowest RMSE (0.7158), while Track0 exhibited the shortest median distance to keypoints (0.0964\,m). Merging Tracks0 and3 combined both accurate prediction and close ground-truth alignment.
\begin{table}[!tb]
    \centering
    \caption{RMSE and median distance to keypoints for each track in a single sample.}
    \renewcommand{\arraystretch}{1.5}
    \small
    \begin{tabular}{|p{2cm}|p{3cm}|p{3cm}|}
    \hline
    \textbf{Track ID} & \textbf{RMSE} & \textbf{Distance (m)} \\
    \hline
    0 & 0.9317 & 0.0964 \\ \hline 
    1 & 1.3630 & 1.4454 \\ \hline 
    2 & 1.0531 & 1.0316 \\ \hline
    3 & 0.7158 & 0.4745 \\
    \hline 
    \end{tabular}
    \label{tab:rmse_distance}
\end{table}

\begin{figure}[h!]
    \centering
    \includegraphics[width=1\textwidth]{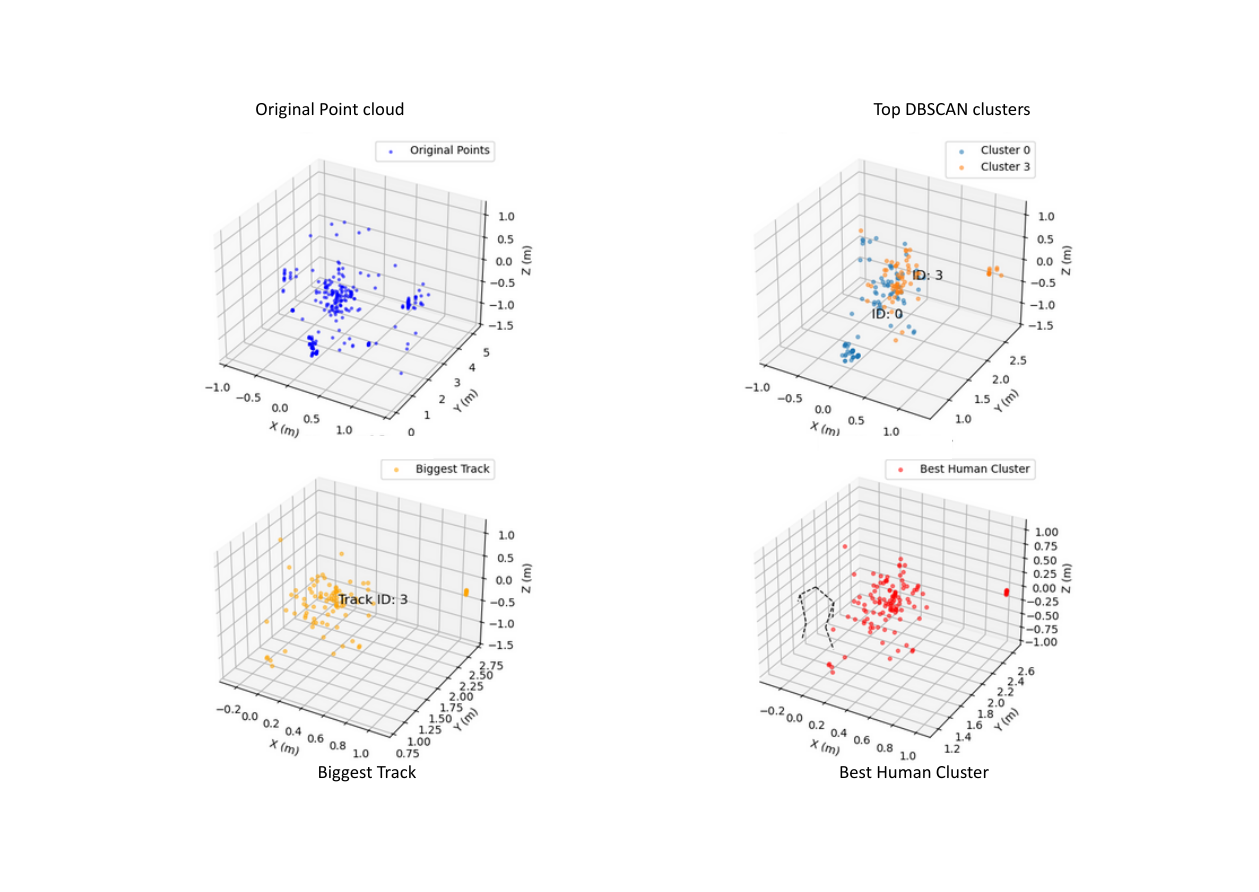}
    \vspace{-30pt} 
    \caption{(Top-left) Original dataset; (top-right) largest DBSCAN clusters; (bottom-left) largest track (Track 3); (bottom-right) best human cluster.}
    \label{fig:largest1}
\end{figure}

Figure \ref{fig:largest1} visualizes the pipeline’s progression from raw point clouds to the final refined cluster. Combining the best metrics reduced each sample from 1100 points to a range of 60--300, retaining motion data while discarding noise.

\subsection{Comparison: Data processing Pipeline vs. Baseline Model Performance}
The results in Table~\ref{tab:method_eval} provide a detailed comparison of various data preprocessing strategies for HAR using mmWave radar data, with a particular focus on their accuracy across four deep learning models (DGCNN, Pointformer, PointNet++, and PointMLP) and their corresponding computational costs. The baseline row shows the performance of each model without any preprocessing applied, serving as a reference point for evaluating the effectiveness of different processing pipelines.
\subsubsection{Individual Methods}
Among the individual preprocessing methods, Kalman filtering (KM) and DBSCAN (DS) both significantly improve model accuracy across the board compared to the baseline. DBSCAN, in particular, demonstrates consistently high performance—with accuracies above 98\% on all four models—highlighting its strength in filtering out noise and isolating meaningful point clusters. Kalman filtering also shows strong performance, but slightly lower than DBSCAN on DGCNN and Pointformer. In contrast, the Hungarian algorithm (HG) alone performs inconsistently. While it provides a noticeable improvement on Pointformer and PointNet++, it underperforms significantly on DGCNN, suggesting that its utility as a standalone method is limited and may be model-dependent. Its relatively high processing time also reduces its attractiveness as a preprocessing step to be used independently.
\subsubsection{Two-Method Combinations}
When combining two methods, the best overall performance is achieved with the Hungarian + DBSCAN (HG + DS) combination. This pairing maintains high accuracy across all models, with very minimal additional computational cost compared to DBSCAN alone. This suggests a synergistic effect where DBSCAN helps clean the data, and the Hungarian algorithm enhances temporal or spatial consistency. Interestingly, combining Kalman and Hungarian (KM + HG) or DBSCAN and Kalman (DS + KM) leads to a drop in accuracy for DGCNN and Pointformer—models that rely more on the geometric structure of the input. This performance degradation may indicate that over-smoothing or misalignment between the clustering and tracking stages could be interfering with the models’ ability to extract meaningful features. In particular, DGCNN, which builds dynamic graphs based on neighborhood relations, may suffer when the preprocessing distorts those local structures.
\subsubsection{Full Pipeline (Three Methods)}
The full pipeline that combines DBSCAN, Kalman filtering, and the Hungarian algorithm (DS + KM + HG) yields strong and balanced performance across all models. While its accuracy slightly trails the best individual or two-method setups in some cases, it maintains high consistency (above 91\% on all models) and delivers robustness across different model architectures. However, the computational cost is significantly higher than any of the simpler combinations (0.2369 seconds), which may limit its applicability in real-time or resource-constrained environments.
\begin{table}[!tbp]
\centering
\caption{Accuracy (\%) and processing time for different data preprocessing methods using MiliPoint dataset with baseline HAR models.}
\scriptsize
\begin{tabular}{|l|c|c|c|c|c|}
\hline
{Methods} & {\makecell{Processing\\time (s)}} & {DGCNN\cite{DGCNN}} & {Pointformer\cite{Pointformer}} & {PointNet++\cite{PointNet++}} & {PointMLP\cite{PointMLP}} \\
\hline
Baseline  & N/A & 13.61 & 29.27 & 34.45 & 18.37 \\
Kalman (KM) & 0.0003 & 98.45 & 96.11 & 98.48 & 98.57 \\
DBSCAN (DS) & 0.0024 & 98.19 & 96.14 & 98.54 & 98.42 \\
Hungarian (HG) & 0.0036 & 72.47 & 95.20 & 98.10 & 97.55 \\
KM + HG & 0.0038 & 42.56 & 44.62 & 52.88 & 47.06 \\
HG + DS & 0.0042 & 98.20 & 95.64 & 98.27 & 98.23 \\
DS + KM & 0.0035 & 43.78 & 43.94 & 51.42  & 47.78 \\
DS + KM + HG & 0.2369 & 95.26 & 91.45 & 95.67 & 95.31 \\
\hline
\end{tabular}
\label{tab:method_eval}
\end{table}

\section{Conclusion}
\label{sec:conclusion}
This study conducted a comprehensive evaluation of three widely used data preprocessing methods—DBSCAN, the Hungarian Algorithm (HA), and Kalman Filtering (KF)—in the context of HAR using mmWave radar data. The results show that DBSCAN is highly effective in isolating human motion from environmental noise, providing a strong standalone improvement in recognition accuracy. The Hungarian Algorithm enhances temporal consistency by maintaining accurate cluster associations across frames, while Kalman Filtering contributes to smoother and more reliable motion tracking through its predictive modeling. When integrated, these methods form a robust preprocessing pipeline that significantly boosts the quality of input data for downstream HAR models. Our analysis demonstrated that combining all three methods achieves the highest overall recognition accuracy—up to 95.67\% with PointNet++—though at the cost of increased processing time. For real-time or resource-limited scenarios, DBSCAN alone or in combination with the Hungarian Algorithm provides an efficient and effective compromise, delivering strong performance with minimal computational cost. These findings highlight the importance of thoughtful method selection and combination in HAR pipelines. They also offer practical guidance for deploying radar-based HAR systems in environments where accuracy, privacy, and efficiency must be carefully balanced. 



\end{document}